\documentclass{article}
\usepackage{amsfonts}
\usepackage{amsmath}
\usepackage{graphicx}
\usepackage{hyperref}
\usepackage{multirow}
\usepackage{mwe}
\usepackage{styles/spconf}
\usepackage{subcaption}
\usepackage[table,xcdraw]{xcolor}
\usepackage{pifont}

\newcommand{\cmark}{\ding{51}}
\newcommand{\xmark}{\ding{55}}

\usepackage[style=ieee, maxbibnames=1,minbibnames=1,sorting=none,doi=false,isbn=false,url=false,eprint=false,maxcitenames=20]{biblatex}
\usepackage{enumitem}
\setlist{nosep, leftmargin=14pt}

\title{Indication as Prior Knowledge for Multimodal Disease Classification in Chest Radiographs with Transformers}

\name{Grzegorz Jacenk\'ow$^{1}$ \qquad Alison Q. O'Neil$^{1, 2}$ \qquad Sotirios A. Tsaftaris$^{1, 3}$}
\address{
  $^{1}$ The University of Edinburgh
  $^{2}$ Canon Medical Research Europe
  $^{3}$ The Alan Turing Institute
}

\begin{document}

\maketitle

\begin{abstract}
When a clinician refers a patient for an imaging exam, they include the reason
(e.g. relevant patient history, suspected disease) in the scan request; this
appears as the indication field in the radiology report. The interpretation and
reporting of the image are substantially influenced by this request text,
steering the radiologist to focus on particular aspects of the image. We use the
indication field to drive better image classification, by taking a transformer
network which is unimodally pre-trained on text (BERT) and fine-tuning it for
multimodal classification of a dual image-text input. We evaluate the method on
the MIMIC-CXR dataset, and present ablation studies to investigate the effect of
the indication field on the classification performance. The experimental results
show our approach achieves 87.8 average micro AUROC, outperforming the
state-of-the-art methods for unimodal (84.4) and multimodal (86.0)
classification. Our code is available at \url{https://github.com/jacenkow/mmbt}.
\end{abstract}


\section{Introduction}
Chest radiography remains the most common imaging examination for the diagnosis
and treatment of a variety of lung conditions such as pneumonia, cancer, and
even COVID-19. Automation of X-ray interpretation could considerably improve
healthcare systems, lowering costs and addressing the pressing challenge of
expert shortage \cite{baltruschat2019comparison}. Yet, current techniques for
clinical decision support mostly focus on a single modality (e.g. patient's
X-ray) and do not take into account complementary information which might be
already available in a hospital's database (e.g. patient's clinical history)
\cite{jacenkow2020inside,jacenkow2019conditioning}. We are particularly
interested in providing the indication field, i.e., the motivation for the
patient's screening examination. This field may include the patient's history, a
request to evaluate a particular condition, and other clues which can steer the
radiologist's attention to particular imaging features. The indication field is
often the only information provided by the referring physician
\cite{obara2015evaluating}, and can influence the interpretation of the imaging
exam \cite{leslie2000influence}. In this paper, we want to design a
vision-and-language model that is able to use such text-based side information
to aid and complement disease classification.

\begin{figure}[t!]
  \centering
  \includegraphics[width=\linewidth]{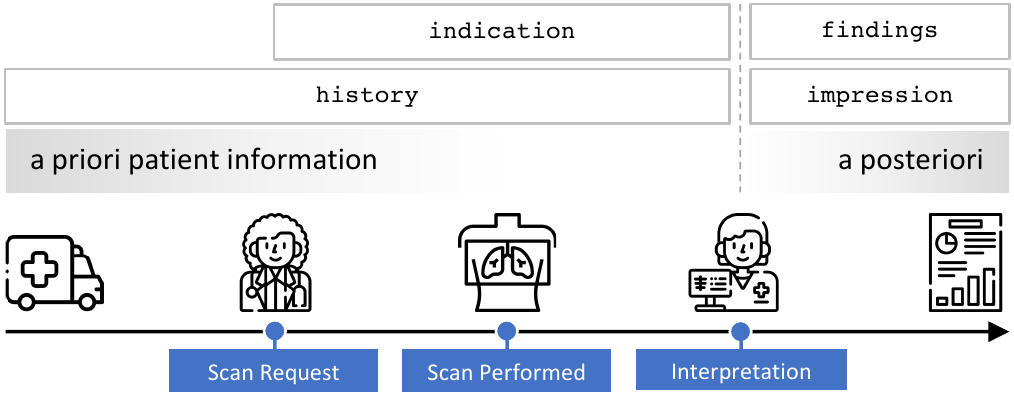}
  \caption{We consider the problem of classifying chest X-ray images given the
           patient information in a free-text form. We only use knowledge about
           the patient collected before the imaging examination and do not require radiologist
           intervention as opposed to most prior studies.}
\end{figure}

Current state-of-the-art methods for vision-and-language tasks (such as
VisualBERT \cite{li2019visualbert}) are mostly based on transformer
architectures, which require extensive pre-training. The process typically
involves using a dataset with annotated bounding boxes around the objects of
interests, such as Conceptual Captions \cite{sharma2018conceptual}, to
initialise the weights, which are later fine-tuned to the final task.
Unfortunately, the biomedical community lacks domain-specific yet general
multimodal datasets which could be used for pre-training large transformer
networks. To address this problem, one could leverage existing unimodal models,
and fine-tune the models to a multimodal task as proposed in multimodal BERT
(MMBT) \cite{kiela2019supervised}, which we evaluate on a biomedical task. As
BERT does not provide the means to process imaging input, MMBT embeds image
features from a ResNet-152 \cite{he2016deep} classifier. 

We evaluate the ability of a unimodally pre-trained BERT model to process
biomedical imaging and non-imaging modalities during the fine-tuning step.
Specifically, we use chest radiographs and the indication field from associated
radiology reports to perform multi-label classification. The network can be
pre-trained on unimodal datasets which are more common than multimodal, but it
is still capable of learning multimodal interactions during the fine-tuning
step.

\noindent
\textbf{Contributions: (1)} We present a strong baseline for multimodal
classification of chest radiographs; \textbf{(2)} We evaluate the model with the
prior work achieving the new state-of-the-art results, and show its robustness
to adversarial changes in text.
\section{Related Work}
\textbf{Chest X-Ray Classification.} Most work for classifying chest
radiographs has been based on existing convolutional neural networks (CNNs)
with ResNet-50 \cite{he2016deep} being the most popular architecture
\cite{baltruschat2019comparison}. Several works have proposed to exploit
non-imaging data such as patient's demographics to improve performance.
The information is often fused before the final classification layer by
concatenating imaging and non-imaging features
\cite{baltruschat2019comparison,li2021lesion}; this late fusion of modalities
limits the methods to model signal-level interactions between imaging and
non-imaging information. Moreover, the non-imaging modality has limited
expressive power as it only relates to basic demographics and not to the
patient's history. We decide to use the indication field from full-text reports.
The free-text input includes relevant information for the imaging procedure,
allowing the network to learn more complex interactions between input images and
the associated reports.

\noindent
\textbf{Learning with Radiology Reports.} TieNet \cite{wang2018tienet}
combines image-text pairs to learn a common embedding for classification and
report generation. The method uses multi-level attention with CNN and RNN
networks for processing radiographs and reports respectively.
However, the full report is expected as input, which requires a radiologist to
render findings first. Recently, two methods
\cite{chauhan2020joint,sylvain2020cross} proposed to leverage information
available in radiology reports to improve performance of image-only
classification. The methods are optimised with a loss encouraging learning a
shared representation between two modalities, while keeping the modalities (and
the downstream tasks) decoupled. The results show improvement in classification
performance, but the methods ignore the additional non-imaging information
during inference. Our work follows the same motivation as
\cite{van2020towards,tian2019towards}, where the methods only include
information available prior to the examination. The first work
\cite{tian2019towards} to include the indication field uses the information only
to improve the quality of rendering the diagnosis (impression field) leaving the
classification head only dependent on the imaging features. The setup was
adapted in \cite{van2020towards} to support classification (and impression
generation) with both modalities. The authors use an attention layer to merge
the output of two feature extractors for image and text, which we term a middle
fusion approach. We propose to use a transformer network which is capable of modelling
the interactions at the word level, enabling the network to perform more complex
fusion. Recently, a study \cite{li2020comparison} has shown the
visual-linguistic BERT models are suitable for processing chest radiographs and
the associated radiology reports, outperforming unimodal approaches for
text-only. However, the evaluated models use full-text reports making the use of
the imaging input negligible and clinically unpractical. By contrast, we propose
information only available to the radiologist prior to developing a report
to drive better image classification rather than labelling text reports which
already can be effortlessly classified by ruled-based approaches.

\section{Methodology}

\begin{figure}[]
  \centering
  \includegraphics[width=\linewidth]{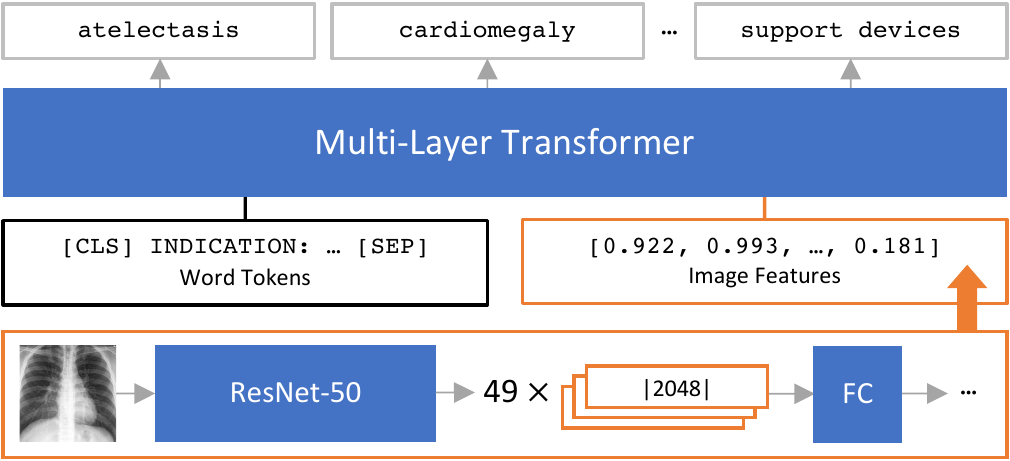}
  \caption{The overview of method. We extend a multi-layer transformer
           pre-trained on textual data with imaging input. The images are
           provided as features extracted from a ResNet-50 network. The features
           are reshaped to 49 vectors of 2048 dimensions each and combined with
           two embeddings describing segment (image or text) and position of the
           token.}
\end{figure}

State-of-the-art methods for modelling vision-language tasks are mostly based on
the transformer architecture where the second segment provides visual tokens from an
image feature extractor. However, pre-training also requires large, and
general multimodal datasets where the visual objects are annotated with bounding
boxes, and such datasets are lacking in the biomedical community. We exploit
unimodally pre-trained BERT model and fine-tune it to a multimodal task.

\noindent
\textbf{Backbone Network:} We adapt BERT \cite{devlin2018bert} as our backbone
network. We use the Hugging Face implementation of
\texttt{bert-base-uncased} pre-trained on textual input. As
the original model has not been developed for visual-linguistic tasks, we learn
a new embedding for the image tokens.

\noindent
\textbf{Image Encoder:} Our method uses ResNet-50 as the image feature
extractor. We first fine-tuned the network pre-trained on ImageNet to classify
chest radiographs (also a baseline method) and removed the last pooling layer.
The network outputs 2048 feature maps of $7 \times 7$, which we reshape to 49
vectors. Our image tokens are the sum of three embeddings, i.e., the linear
projection of the $i$\textsuperscript{th} vector ($i \in [1, 49]$), the position
of the vector $i$, and the segment indicating the imaging modality. We keep the
weights of the image encoder unfrozen during the fine-tuning step of the whole
model.

\noindent
\textbf{Classification Head:} We use the final representation of \texttt{[CLS]}
token to fine-tune our model for classification. We apply a multi-layer perceptron
$\{768 - 768 - 14\}$ with GELU activation functions
and layer normalisation. The last layer applies a \textit{sigmoid} function to
each of fourteen nodes. 

\noindent
\textbf{Loss Function:} We optimise a binary cross-entropy loss with class
weighting that is inversely proportional to the number of examples in the
training set.

\begin{table*}[ht!]
  \begin{center}
    \caption{Quantitative results on the MIMIC-CXR dataset. We report average
             accuracy, precision, recall, $F_1$ score, and the area under the ROC (AUROC).
             The results are reported as average over three runs with standard deviation
             reported as subscript. The number in \textbf{bold} denotes the best performance
             within the metric.}
    \label{tab:overall}
    \resizebox{\linewidth}{!}{%
    \begin{tabular}{l|c|c||c||c|c||c|c||c|c||c|c}
      \multirow{2}{*}{Method} & \multicolumn{2}{c||}{Modality} & \multirow{2}{*}{Accuracy} & \multicolumn{2}{c||}{Precision} & \multicolumn{2}{c||}{Recall} & \multicolumn{2}{c||}{$F_1$} & \multicolumn{2}{c}{AUROC} \\
      & Image & Text && Macro & Micro & Macro & Micro & Macro & Micro & Macro & Micro \\
        \hline
        CheXpert Labeler  & \xmark & \cmark & 80.6 & 9.3 & 13.4 & 18.8 & 27.0 & 8.5 & 17.9 & 51.13 & 53.3 \\
        BERT         & \xmark & \cmark & $85.1_{\pm 0.2}$ & $21.6_{\pm 1.0}$ & $32.9_{\pm 0.9}$ & $47.2_{\pm 5.8}$ & $54.7_{\pm 0.9}$ & $26.1_{\pm 1.1}$ & $41.1_{\pm 0.9}$ & $71.1_{\pm 1.2}$ & $81.7_{\pm 0.8}$ \\
        ResNet-50 & \cmark & \xmark & $86.0_{\pm 0.2}$ & $26.0_{\pm 1.1}$ & $43.7_{\pm 2.1}$ & $34.0_{\pm 1.8}$ & $57.4_{\pm 1.5}$ & $27.4_{\pm 0.4}$ & $49.5_{\pm 0.8}$ & $73.8_{\pm 0.5}$ & $84.4_{\pm 0.6}$ \\
        \hline
        Attentive         & \cmark & \cmark & $86.8_{\pm 0.1}$ & $26.8_{\pm 0.5}$ & $44.2_{\pm 0.7}$ & $34.7_{\pm 0.2}$ & $61.3_{\pm 0.6}$ & $29.1_{\pm 0.4}$ & $51.4_{\pm 0.3}$ & $76.6_{\pm 0.3}$ & $86.0_{\pm 0.2}$ \\
        MMBT & \cmark & \cmark & $\textbf{87.7}_{\pm 0.2}$ & $\textbf{30.8}_{\pm 0.3}$ & $\textbf{47.8}_{\pm 0.7}$ & $\textbf{55.4}_{\pm 1.8}$ & $\textbf{64.7}_{\pm 0.7}$ & $\textbf{35.0}_{\pm 0.6}$ & $\textbf{55.0}_{\pm 0.6}$ & $\textbf{80.6}_{\pm 0.1}$ & $\textbf{87.8}_{\pm 0.1}$ \\
    \end{tabular}%
    }
  \end{center}
\end{table*}

\section{Experiments}
\subsection{Dataset}
We use the MIMIC-CXR dataset
\cite{johnson2019mimic,johnson2019physionet,goldberger2000physiobank}, which
consists of 377,110 chest X-ray images associated with 227,835 post-screening
reports of 65,379 patients taken at the Beth Israel Deaconess Medical Center
Emergency Department. We limit the experiments to examinations with frontal
images (AP/PA), and reports with \texttt{indication} or \texttt{history} fields
explicitly. Our final evaluation is based on 210,538 studies following the
official splits between training (205,923), validation (1695) and test sets
(2920).

\noindent
\textbf{Labelling.} The original data are not labelled for the classification
task. We use the CheXpert Labeler \cite{irvin2019chexpert} to extract fourteen
labels from full radiology reports: atelectasis, cardiomegaly, consolidation,
edema, enlarged cardiomediastinum, fracture, lung lesion, lung opacity, no
finding, pleural effusion, pleural (other), pneumonia, pneumothorax, and support
devices. We set the task as a multilabel problem with positive-vs-rest
classification\footnote{CheXpert Labeler is capable of assigning each label one
of four values - positive, negative, uncertain and no mention. We only select
the positive instances.}.

\noindent
\textbf{Pre-processing.} The images were taken from the MIMIC-CXR-JPG
dataset and resized to $224 \times 224$ pixels. We normalise the images to zero
mean and unit of standard deviation. The text input has been stripped from
special characters (e.g. "\_\_", "\textbackslash") and all characters converted
to lower case.

\subsection{Baselines}
We compare the investigated method to several baselines:

\begin{itemize}
  \item \textbf{CheXpert Labeler \cite{irvin2019chexpert}:} This is the
        rules-based method used to extract the original fourteen labels from the
        full reports. We apply this method to the indication fields.
  \item \textbf{BERT \cite{devlin2018bert}:} We use the unimodal BERT network
        which is the backbone of the proposed method with no access to the
        imaging input. We use the same classification head to fine-tune the
        network for classification.
  \item \textbf{ResNet-50 \cite{he2016deep}:} We use the ResNet-50 network
        pre-trained on ImageNet (image feature extractor in the proposed
        method), which we fine-tune to classify the chest radiographs.
  \item \textbf{Attentive \cite{van2020towards}:} We compare our model
        to the multimodal approach presented in \cite{van2020towards}.
        The method uses ResNet-50 and BioWordVec \cite{zhang2019biowordvec}
        with GRU units for feature extraction, with the two branches merged
        using an attention layer. The original method also
        generates impression fields (not included in our pipeline).
\end{itemize}

\subsection{Experimental Setup}
All baseline methods and the proposed technique were implemented with the
multimodal framework (MMF) \cite{singh2020mmf}. We train the models for 14
epochs with a batch size of 128. We use the Adam optimiser
with weight decay (0.01). We set the learning rate to $5 \times 10^{-5}$ with a
linear warm-up schedule for the first 2000 steps. We apply the early stopping
criterion of multi-label micro $F_1$ score evaluated on the validation set. We
repeat each experiment three times with different seeds to account for variance
due to random weight initialisation.

\begin{table*}[h!]
  \begin{center}
  \caption{The performance of the MMBT to robustness evaluation and
           manipulation to the indication field. We use the evaluation scheme
           proposed in \cite{araujo2020adversarial} and further extend with
           swapping the indication field (no input, stop words, different patient).}
  \label{tab:attacks}
    \resizebox{0.90\linewidth}{!}{%
    \begin{tabular}{l||c||c|c||c|c||c|c||c|c}
      \rowcolor[HTML]{FFFFFF}
      & & \multicolumn{2}{c||}{Precision} & \multicolumn{2}{c||}{Recall} & \multicolumn{2}{c||}{$F_1$} & \multicolumn{2}{c}{AUROC} \\
      \rowcolor[HTML]{FFFFFF}
      \multirow{-2}{*}{Robustness Evaluation} & \multirow{-2}{*}{Accuracy} & Macro & Micro & Macro & Micro & Macro & Micro & Macro & Micro \\
      \hline
      Baseline & ${87.7}_{\pm 0.2}$ & ${30.8}_{\pm 0.3}$ & ${47.8}_{\pm 0.7}$ & ${55.4}_{\pm 1.8}$ & ${64.7}_{\pm 0.7}$ & ${35.0}_{\pm 0.6}$ & ${55.0}_{\pm 0.6}$ & ${80.6}_{\pm 0.1}$ & ${87.8}_{\pm 0.1}$ \\
      \hline
      Character Swap & $87.0_{\pm 0.2}$ & $27.4_{\pm 0.5}$ & $44.7_{\pm 0.7}$ & $48.3_{\pm 8.0}$ & $62.1_{\pm 0.6}$ & $30.4_{\pm 0.7}$ & $52.0_{\pm 0.6}$ & $78.0_{\pm 0.4}$ & $86.5_{\pm 0.1}$ \\
      Keyboard Typo & $86.9_{\pm 0.2}$ & $27.6_{\pm 0.4}$ & $45.3_{\pm 0.6}$ & $46.6_{\pm 3.5}$ & $61.8_{\pm 1.0}$ & $30.4_{\pm 0.1}$ & $52.3_{\pm 0.6}$ & $78.2_{\pm 0.1}$ & $86.4_{\pm 0.1}$ \\
      Synonyms & $87.2_{\pm 0.2}$ & $29.1_{\pm 0.8}$ & $46.1_{\pm 0.3}$ & $49.0_{\pm 1.5}$ & $62.6_{\pm 0.8}$ & $32.7_{\pm 0.4}$ & $53.1_{\pm 0.5}$ & $79.3_{\pm 0.3}$ & $86.9_{\pm 0.1}$ \\
      \hline
      Missing Field & $86.2_{\pm 0.2}$ & $24.4_{\pm 1.6}$ & $42.0_{\pm 1.4}$ & $38.3_{\pm 3.1}$ & $58.9_{\pm 1.2}$ & $27.0_{\pm 1.2}$ & $49.0_{\pm 0.8}$ & $75.1_{\pm 0.3}$ & $84.6_{\pm 0.3}$ \\
      Stop Words Noise & $86.2_{\pm 0.1}$ & $20.2_{\pm 0.7}$ & $35.0_{\pm 1.6}$ & $38.6_{\pm 4.9}$ & $60.7_{\pm 0.4}$ & $24.0_{\pm 0.7}$ & $44.4_{\pm 1.3}$ & $74.7_{\pm 0.3}$ & $84.3_{\pm 0.2}$ \\
      Indication Swap & $84.1_{\pm 0.1}$ & $19.7_{\pm 0.8}$ & $33.4_{\pm 0.6}$ & $30.1_{\pm 0.6}$ & $49.3_{\pm 0.5}$ & $22.7_{\pm 0.8}$ & $39.8_{\pm 0.6}$ & $69.1_{\pm 0.4}$ & $80.0_{\pm 0.4}$ \\
    \end{tabular}%
    }
  \end{center}
\end{table*}

\subsection{Results: Classification Performance} We report the performance of
the tested methods using label-wise accuracy, precision, and recall metrics
where we consider a separate classifier for each of fourteen classes. The
overall quantitative results are shown in Table \ref{tab:overall}. We observe
the CheXpert Labeler has the weakest performance across all of the reported
metrics. The method is a rule-based approach, so it cannot learn associations
between the content of indication fields and the labels, but will pick up only
explicit mentions. This problem is mitigated by BERT (text-only) classifier
which outperforms the labeler in all metrics (+53.3\% improvement in micro
AUROC). We further notice the image-only based classifier (ResNet-50)
outperforms the BERT in all metrics except recall (macro) with micro AUROC
improved by +3.3\%. These findings are consistent with our expectation images
contain the investigation results requested to help determine a diagnosis,
compared to the text modality which describes only the clinician's suspicion
based on patient information prior to imaging. The Attentive
\cite{van2020towards} baseline, which uses both image and text, outperforms the
image- and text-only methods in all reported metrics with micro AUROC improved
by 1.9\% comparing to the best unimodal baseline. Finally, the multimodal BERT
outperforms all unimodal and multimodal baselines with 2\% margin. The method
relies on the early fusion approach (as opposed to middle fusion in Attentive)
enabling the network to learn correlation and interactions between the
modalities with low-level features. Moreover, we present per-class performance
in Table \ref{tab:auroc}, where the investigated method consistently outperforms
the baselines in each of the fourteen classes.

\begin{table}[]
  \centering
  \caption{AUROC results per-class for the tested methods.}
  \resizebox{\linewidth}{!}{%
    \begin{tabular}{l||c|c||c||c}
      \rowcolor[HTML]{FFFFFF}
      AUROC & ResNet-50 & BERT & Attentive & MMBT \\
      \hline
      \rowcolor[HTML]{E7E7E7}
      Atelectasis     & $72.6_{\pm 1.4}$ & $66.5_{\pm 0.6}$ & $73.6_{\pm 0.3}$ & $\textbf{75.8}_{\pm 0.2}$ \\
      Cardiomegaly    & $74.6_{\pm 0.9}$ & $74.1_{\pm 0.4}$ & $77.3_{\pm 0.6}$ & $\textbf{82.6}_{\pm 0.2}$ \\
      \rowcolor[HTML]{E7E7E7}
      Consolidation   & $69.9_{\pm 0.9}$ & $66.8_{\pm 0.9}$ & $73.4_{\pm 0.6}$ & $\textbf{77.1}_{\pm 0.3}$ \\
      Edema           & $81.6_{\pm 0.5}$ & $70.7_{\pm 0.4}$ & $81.9_{\pm 0.5}$ & $\textbf{84.3}_{\pm 0.4}$ \\
      \rowcolor[HTML]{E7E7E7}
      Enlarged Card.  & $63.3_{\pm 0.9}$ & $69.0_{\pm 0.5}$ & $66.6_{\pm 1.5}$ & $\textbf{74.3}_{\pm 1.1}$ \\
      Fracture        & $63.3_{\pm 0.9}$ & $65.9_{\pm 1.5}$ & $67.0_{\pm 0.7}$ & $\textbf{72.9}_{\pm 2.0}$ \\
      \rowcolor[HTML]{E7E7E7}
      Lung Lesion     & $66.5_{\pm 2.5}$ & $69.9_{\pm 1.8}$ & $70.9_{\pm 0.7}$ & $\textbf{75.9}_{\pm 1.2}$ \\
      Lung Opacity    & $68.3_{\pm 0.9}$ & $62.5_{\pm 0.7}$ & $69.2_{\pm 0.2}$ & $\textbf{71.5}_{\pm 0.5}$ \\
      \rowcolor[HTML]{E7E7E7}
      No Finding      & $77.9_{\pm 0.8}$ & $72.6_{\pm 0.9}$ & $80.9_{\pm 0.2}$ & $\textbf{83.1}_{\pm 0.3}$ \\
      Ple.   Effusion & $86.1_{\pm 0.6}$ & $71.6_{\pm 0.4}$ & $87.1_{\pm 0.1}$ & $\textbf{88.6}_{\pm 0.1}$ \\
      \rowcolor[HTML]{E7E7E7}
      Pleural Other   & $79.0_{\pm 1.8}$ & $72.3_{\pm 1.2}$ & $78.6_{\pm 1.0}$ & $\textbf{86.9}_{\pm 0.5}$ \\
      Pneumonia       & $66.1_{\pm 2.1}$ & $68.9_{\pm 0.4}$ & $70.9_{\pm 0.8}$ & $\textbf{75.2}_{\pm 0.7}$ \\
      \rowcolor[HTML]{E7E7E7}
      Pneumothorax    & $78.3_{\pm 0.4}$ & $86.5_{\pm 1.0}$ & $85.1_{\pm 0.7}$ & $\textbf{88.0}_{\pm 0.3}$ \\
      Support Devices & $85.9_{\pm 0.5}$ & $88.7_{\pm 0.3}$ & $90.5_{\pm 0.2}$ & $\textbf{92.2}_{\pm 0.1}$ \\
    \end{tabular}
  }
  \label{tab:auroc}
\end{table}

\subsection{Results: Robustness to Textual Input}
Overburdened clinicians may introduce or propagate typographical errors
while composing a request for imaging examination. We argue it is essential to
evaluate models along with the main performance metrics on robustness to changes
of the textual input such as common mistakes and use of synonyms. To achieve this goal,
we test the MMBT model to textual changes with an evaluation scheme proposed in
\cite{araujo2020adversarial} which we further extended. We mimic a human
operator who commits typographical errors and expresses the original medical terms
with synonyms. We only select biomedical terms to proceed with the following
word/sentence manipulation:

\begin{itemize}
  \item \textbf{Character Swap}: swapping two consecutive characters at random,
        e.g. fever $\rightarrow$ fevre.
  \item \textbf{Keyboard Typo:} selecting a random character and replacing with an
        adjacent one, e.g. fever $\rightarrow$ f3ver.
  \item \textbf{Synonyms:} selecting a synonym for a given biomedical term using
        the UMLS database, e.g. fever $\rightarrow$ pyrexia.
  \item \textbf{Missing Field/Stop Words Noise:} replacing the indication field
        with an empty string or a sentence using only stop words.
  \item \textbf{Indication Swap:} selecting a random indication from another
        patient ensuring no single positive class is shared between two
        patients.
\end{itemize}

\noindent
The results are presented in Table \ref{tab:attacks}. The tested method is
resistant to common typographical errors and capable of processing synonyms
affecting the performance at most by -1.7\% micro AUROC (keyboard typo). When
the method does not have access to the corresponding indication fields, the
performance of the multimodal transformer is on par with ResNet-50 (micro
AUROC). The experiment has shown the method improves while the patient's history
is provided, yet is still capable of processing only images with no textual
input, a common scenario in emergency departments. However, replacing the
original indication field with a different patient significantly affects the
performance (-16.6 \% and -9.8\% on macro and micro AUROC, respectively). The
test has the most notable effect expected on the method (providing
clues conflicting with the imaging input), proving that the model uses both
modalities to render a decision.

\section{Conclusion}
We evaluated a unimodally pre-trained BERT model on multimodal chest radiograph
classification supported by the indication field. We extended the BERT model
with an image feature extractor and show it can successfully learn imaging
modality, beating the previous state-of-the-art approaches for this task (+4\%
and +2\% micro AUROC for uni- and multimodal baselines, respectively). These
promising results show the model can leverage prior knowledge about the patient
for a more accurate image diagnosis. We presented the model as resistant to
typographical errors, capable of handling synonyms, and missing text input
matching image-only baseline.

\section{Compliance with Ethical Standards}
This research study was conducted retrospectively using human subject data made
available in open access. Ethical approval was not required as confirmed by the
license attached with the open access data.

\section*{Acknowledgements}
This work was supported by the Engineering and Physical Sciences Research
Council [grant number EP/R513209/1] and Canon Medical Research Europe. S.A.
Tsaftaris acknowledges the support of the Royal Academy of Engineering and the
Research Chairs and Senior Research Fellowships scheme [grant number
RCSRF1819\textbackslash8\textbackslash25].

\printbibliography

\end{document}